\documentclass[letterpaper]{article} 
\usepackage[preprint]{aaai2027}
\usepackage[hyphens]{url}  
\usepackage{graphicx} 
\usepackage{natbib}  
\usepackage{caption} 
\usepackage{amsmath}

\usepackage{booktabs}

\newcommand{\papertitle}{AgentRewind: Recoverable Execution\\ for Long-Horizon LLM Agents}

\newcommand{\paperauthors}{%
  Yu Zhuang\textsuperscript{\rm 1,\rm 3,\rm 4}\equalcontrib,
  Kefei Chen\textsuperscript{\rm 2,\rm 3}\equalcontrib,
  Yitong Duan\textsuperscript{\rm 3}\corresponding,
  Shuxin Zheng\textsuperscript{\rm 3},
  Jian Li\textsuperscript{\rm 2},
  Xu-Yao Zhang\textsuperscript{\rm 1,\rm 4}\corresponding
}

\newcommand{\paperaffiliations}{%
  \textsuperscript{\rm 1}University of Chinese Academy of Sciences\\
  \textsuperscript{\rm 2}IIIS, Tsinghua University\\
  \textsuperscript{\rm 3}Zhongguancun Academy, Beijing, China\\
  \textsuperscript{\rm 4}Institute of Automation, Chinese Academy of Sciences\\
  zhuangyu24@mails.ucas.ac.cn, ckf25@mails.tsinghua.edu.cn, duanyitong@zgci.ac.cn, xyz@nlpr.ia.ac.cn
}

\title{\papertitle}
\author{\paperauthors}
\affiliations{\paperaffiliations}

\usepackage{listings}
\newcommand{\bench}{MettleBench}
\newcommand{\method}{AgentRewind}
\newcommand{\mainpaper}{the main paper}
\newcommand{\maintab}[1]{Table~#1 of \mainpaper}
\newcommand{\mainfig}[1]{Figure~#1 of \mainpaper}

\begin{document}
\maketitle

\begin{abstract}
Many real-world tasks require LLM agents to interact with their environments over long execution horizons. Errors that occur early in execution may propagate through both the agent context and environment state, and their effects may be difficult to reverse through subsequent actions. Existing methods mainly seek to reduce such errors through plan refinement and safety checks but provide little support after errors occur. To enable recovery during long-horizon execution, we present \textbf{AgentRewind}, a runtime recovery framework that records aligned checkpoints of the agent context and controlled environment, allowing agents to return to an earlier state and resume execution with information from previous attempts. We also construct \textbf{MettleBench}, a benchmark for evaluating task
completion and partial progress on long-horizon engineering assignments
containing a series of related requirements. Experiments across tasks, multiple models, execution strategies, and agent harnesses show that AgentRewind improves task success rate and average checklist progress over the compared baselines.
\end{abstract}

\begin{links}
    \link{Code}{https://github.com/Futuresis/replay-agent-recorder}
    \link{Dataset}{https://github.com/Kelvin-Coffee/MettleBench}
\end{links}

\section{Introduction}
\label{sec:introduction}

LLM agents solve tasks through tool-mediated interaction with their
environments. Unlike non-agentic LLMs, they observe environmental states, invoke tools, and adjust subsequent actions based on feedback \citep{react,liu2024agentbench}. They are increasingly applied to long-horizon tasks such as iterative scientific experimentation, cross-application workflows, and repository-level software engineering
\citep{kon2026expbench,li2026windowsworld,swebench}. Improving their
performance on such tasks is therefore an important research objective.

However, operating over long horizons presents distinct challenges for LLM agents. For example, an agent may formulate an incorrect plan early in execution, causing subsequent actions to proceed in the wrong direction \citep{valmeekam2023planbench,liu2024agentbench,wang2026long}. During execution, an agent may also delete critical files, corrupt configurations, or contaminate database states, forcing subsequent actions to proceed in a degraded environment and potentially preventing task completion \citep{toolemu,trivedi2024appworld}.
 Because long-horizon tasks involve more execution steps, agents are more likely to make such errors along the execution trajectory and therefore face a higher risk of task failure.

Existing approaches primarily seek to improve agent performance on long-horizon tasks through plan refinement and safety checks. One line of work generates and revises plans before execution to identify more effective execution paths \citep{planact,lats}. Another line of work applies safety checks during execution to detect unsafe behavior \citep{wang2026agentspec,guardagent}. However, even with a low probability of error at each step, cumulative failure risk grows with trajectory length. These methods do not adequately address recovery after an error has disrupted execution. 

We therefore introduce AgentRewind, a runtime recovery framework that allows agents to rewind execution to an earlier state. When the agent determines that it is unlikely to make further progress along the current trajectory, it can select an earlier checkpoint based on the recorded trajectory. AgentRewind then restores the agent context and environment to the states recorded at that checkpoint, allowing the agent to continue execution from there. It also retains a summary of the previous attempt as rewind memory to guide the agent's subsequent decisions. Through these capabilities, AgentRewind enables agents to recover from errors during task execution, improving their performance on long-horizon tasks. Our main contributions are summarized as follows:
\begin{itemize}
\item We develop AgentRewind, a system that can return both the agent context and the environment state to a selected checkpoint. This allows the agent to continue from an earlier checkpoint and explore a different trajectory when errors prevent further progress.
\item We construct MettleBench, a benchmark derived from real-world
engineering resources, to evaluate task completion and checklist
progress on assignments containing a series of related requirements.
\item We evaluate AgentRewind across multiple base models, execution strategies, and agent harnesses, and show that it improves task success rate and average checklist progress in long-horizon settings.

\end{itemize}

\section{Related Work}
\label{sec:related_work}

\subsection{Long-Horizon Agents}

Long-horizon agents have been studied across embodied, scientific, web-based, software engineering, and database settings. ALFWorld, ScienceWorld, and WebShop evaluate long-horizon interaction in embodied, scientific, and web environments \citep{alfworld,scienceworld,webshop}. InterCode and SWE-bench study interactive programming and repository-level software engineering \citep{intercode,swebench,sweagent}, while Spider 2.0, CoSQL, and BIRD-INTERACT focus on multi-turn database tasks \citep{spider2,cosql,birdinteract}. More recent benchmarks, including TheAgentCompany and Odysseys, extend evaluation to realistic computer-use tasks and show that even strong models often fail to complete long-horizon tasks \citep{theagentcompany,odysseys}.

\subsection{Agent Reliability}

Prior work on agent reliability improves execution through better action generation and safety monitoring. Methods such as ReAct, SayCan, and LATS guide agent behavior using reasoning, planning, search, and environment models \citep{react,saycan,innermonologue,planact,lats,tree_search_agents,web_world_model}. Safety-oriented methods monitor agent execution and identify risky behavior \citep{wang2026agentspec,guardagent,liu2026agentdog}.

Another line of work uses information from previous attempts to improve later behavior. Self-Refine, Reflexion, and ExpeL use feedback from prior attempts to revise subsequent decisions \citep{selfrefine,reflexion,expel}. Related failure-analysis methods diagnose the causes of failure from
completed execution trajectories \citep{agentdebug,agentrx}. These methods use information from previous executions to improve later behavior, but provide little support for recovery after an error occurs during execution.

\subsection{Rollback for LLM Agents}

Several studies have explored rollback for LLM agents. GA-Rollback \citep{ga_rollback} performs stepwise recovery in replayable environments, whereas WebRollback \citep{webrollback}
specializes rollback to browser navigation. DART \citep{dart} selects semantically valid restore points for structured tool agents but relies on explicit control flow and recovery boundaries, limiting its applicability to open-ended agent trajectories.

System-level checkpointing tools such as DMTCP and CRIU provide general mechanisms for restoring process state \citep{dmtcp,criu}. DeltaBox extends these mechanisms to agent sandboxes, enabling efficient rollback of filesystem and process state \citep{deltabox}. However, these systems primarily provide low-level rollback mechanisms rather than an agent-level recovery process for long-horizon execution. AgentRewind combines coordinated restoration of the agent context
and environment state with rewind memory, allowing the agent to return to an earlier checkpoint and continue execution with information from previous attempts.

\section{AgentRewind Framework}
\label{sec:method}

\begin{figure*}[t]
    \centering
    \includegraphics[width=0.95\textwidth]{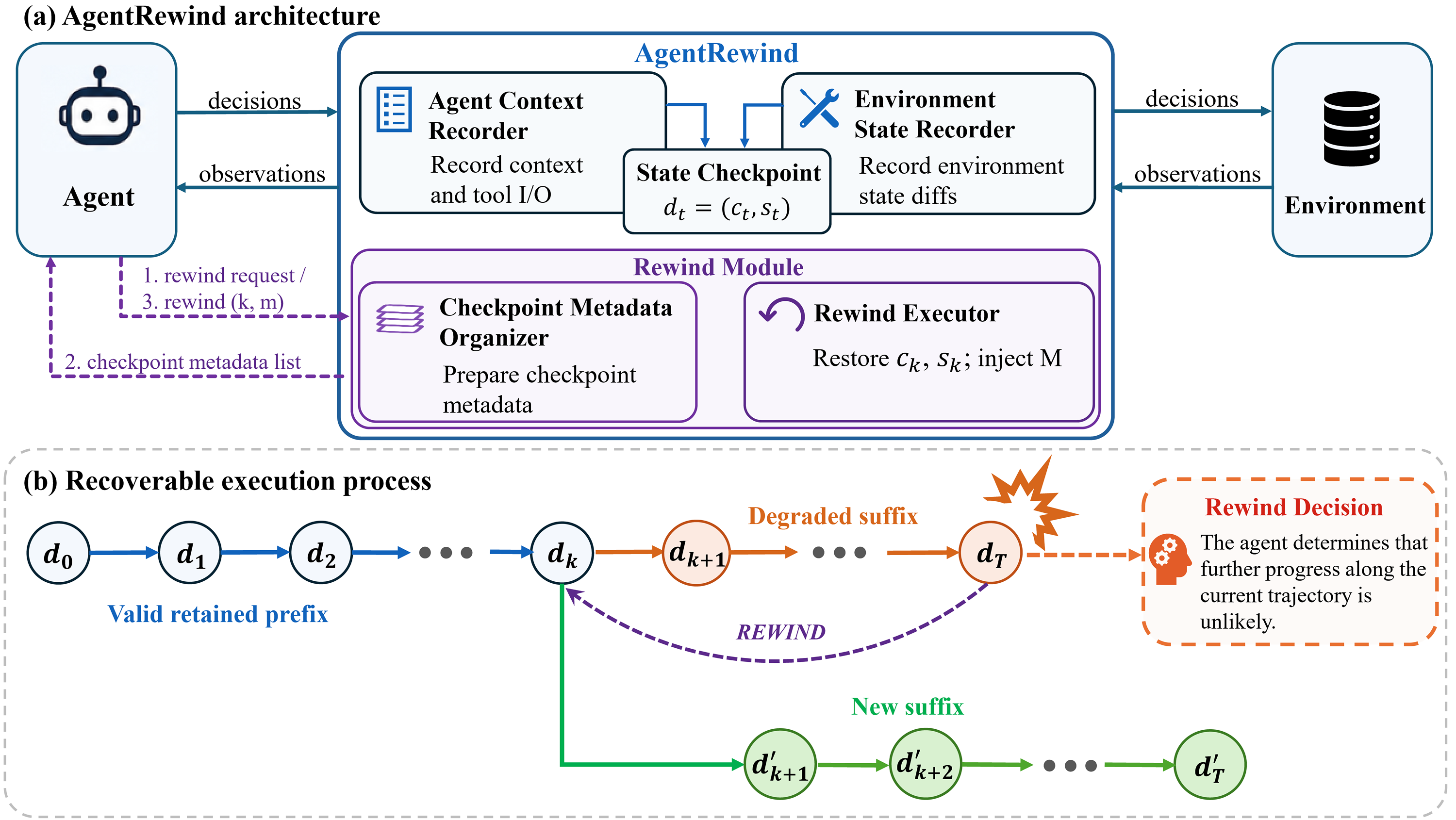}
    \caption{
            Overview of AgentRewind.
            (a) AgentRewind operates as a runtime layer between the agent and the controlled external environment. The context recorder and environment-state recorder create aligned checkpoints \(d_t=(c_t,s_t)\). During rewind, the agent selects a target checkpoint, after which the rewind executor restores the corresponding context and environment state and injects rewind memory.
            (b) An illustrative recoverable execution. When the agent determines that the current trajectory can no longer make progress, it invokes rewind. The trajectory returns to \(d_k\), retains the prefix through that checkpoint, and continues with a new suffix.
            }
\label{fig:agentrewind}
\end{figure*}

AgentRewind provides a runtime recovery framework for long-horizon agent execution. It records recoverable checkpoints that capture the agent context and controlled environment state. When a rewind is triggered, AgentRewind restores the selected checkpoint and injects agent-generated rewind memory into the restored context. Figure~\ref{fig:agentrewind} provides an overview of the system architecture and the recoverable execution process.

\subsection{Long-Horizon Execution with Rewind}
\label{sec:recoverable_execution}

We consider an agent executing long-horizon tasks in a controlled external environment. Given a task instruction \(x\), the agent iteratively changes the environment state through tool calls. At step \(t\), the LLM generates a decision
\(u_t\) from its context \(c_t\). The environment transitions from
\(s_t\) to \(s_{t+1}\) and returns observation \(o_{t+1}\), after which the agent context is
updated:
\begin{equation}
\begin{aligned}
u_t &\sim \pi(c_t),\\
(s_{t+1},o_{t+1}) &= \mathcal{T}(s_t,u_t),\\
c_{t+1} &= \mathcal{U}(c_t,u_t,o_{t+1}).
\end{aligned}
\label{eq:forward-interaction}
\end{equation}
These updates produce a forward-growing trajectory:
\begin{equation}
\tau =
\bigl((c_0,s_0),u_0,o_1,(c_1,s_1),\ldots,(c_T,s_T)\bigr).
\label{eq:forward-trajectory}
\end{equation}

Under standard execution, the trajectory can only grow forward. Once an early decision \(u_k\) introduces an error, its effects may propagate through subsequent contexts \(c_{k+1}\) and environment states \(s_{k+1}\). Even if the agent later recognizes the error, it can only append corrective actions to the existing trajectory and may be unable to fully reverse the state changes caused by the erroneous prefix. 

AgentRewind records the execution state at each LLM decision boundary as a recoverable checkpoint \(d_t\):
\begin{equation}
d_t=(c_t,s_t).
\label{eq:checkpoint}
\end{equation}
For each state checkpoint, AgentRewind constructs associated checkpoint metadata \(\eta_t\). The metadata describes the execution segment between \(d_t\) and \(d_{t+1}\). It is used to present possible rewind locations to the agent. 

Given the trajectory \(\tau_{0:T}\), when the agent determines that the current trajectory can no longer make progress, it can select a historical checkpoint \(d_k\) based on its context and checkpoint metadata list, where \(k \leq T\), and generate rewind memory \(m\) from the current trajectory. AgentRewind adds \(m\) to the set of historical rewind memories \(M\):
\begin{equation}
M \leftarrow M \cup \{m\}.
\label{eq:memory-update}
\end{equation}
AgentRewind then restores the controlled external environment state to \(s_k\) and injects the accumulated rewind memories into the recorded agent context \(c_k\):
\begin{equation}
\begin{aligned}
s'_k &\leftarrow s_k,\\
c'_k &\leftarrow \operatorname{Inject}(c_k,M).
\end{aligned}
\label{eq:aligned-restoration}
\end{equation}
The agent subsequently resumes execution from \((c'_k, s'_k)\), producing a new continuation of the trajectory:
\begin{equation}
\tau' =
\bigl((c'_k,s'_k),u'_k,o'_{k+1},\ldots\bigr).
\label{eq:rewound-trajectory}
\end{equation}
Figure~\ref{fig:agentrewind}(b) illustrates this recoverable execution process. Rather than continuing only from the latest execution state, the agent can return to an earlier checkpoint and resume execution, while retaining information from previous attempts.

\subsection{AgentRewind Runtime Recovery Layer}

\label{sec:recoverable_execution_layer}

Based on the formalization above, AgentRewind augments the standard agent--environment interaction loop with runtime recovery. As shown in Figure~\ref{fig:agentrewind}(a), AgentRewind operates between the agent and the controlled external environment. During normal execution, AgentRewind transparently records the agent's trajectory without altering its interaction with the environment. The recorded trajectory includes the task instruction, LLM inputs and outputs, tool calls, and tool results. It also tracks tool-induced changes to the controlled environment and aligns them with the corresponding context records to form recoverable checkpoints. Therefore, the recorded trajectory supports the joint recovery of the agent context and the controlled environment state.

\subsection{Rewind Module}
\label{sec:rewind_module}

AgentRewind exposes rewind as an additional control action to the agent.
When the agent decides to invoke rewind, AgentRewind organizes the recorded trajectory into checkpoint metadata entries. Each entry summarizes an LLM output together with the tool interactions and environment changes that occur before the next LLM output.
Based on the current execution trajectory, checkpoint metadata, environment feedback, and accumulated rewind memory, the agent selects a checkpoint and generates a new rewind memory that summarizes useful information from the current trajectory, such as falsified hypotheses and alternative strategies.

During rewind execution, AgentRewind terminates the current execution and restores both the external environment and the agent context to the selected checkpoint. The preceding LLM outputs and tool results are restored from the execution log rather than regenerated or re-executed. AgentRewind then injects the accumulated rewind memory into the restored context. The agent then begins generating a new suffix from the restored state.

\subsection{External Environment Recovery Boundary}
\label{sec:environment_recovery_boundary}

The controlled environment restored by AgentRewind is the workspace
directory tree, whether the agent runs locally or in an isolated
container. At each checkpoint, AgentRewind records file-level changes;
rewinding reverts later modifications, restores deleted files, and
removes newly created files. Effects outside the workspace filesystem,
such as network requests, external-service calls, and external runtime
state, cannot be undone. Because the retained prefix is restored from
the execution log, these effects are not
triggered again.

\section{MettleBench}

We next introduce the benchmark used for our main evaluation of
long-horizon execution.
Complex, long-horizon engineering assignments are rarely defined by a single requirement. Instead, they often contain a series of related requirements that together determine whether the assignment has been completed. For example, migrating an internal service may require upgrading its dependencies, transferring configuration and data, preserving compatibility with existing clients, updating deployment scripts, verifying service availability, and documenting the resulting changes. By contrast, many existing agent benchmarks formulate each task around one primary requirement, even when completing it involves multiple execution steps. MettleBench represents this form of long-horizon work by treating each task as a single engineering assignment containing a series of complex requirements. Formally, a task and its ordered acceptance criteria are represented as
\begin{equation}
\mathcal{T}=(x,s_0,\mathcal{U},G),
\qquad
G=(g_1,\ldots,g_n),
\label{eq:mettlebench-task}
\end{equation}
where \(x\) is a natural-language task instruction, \(s_0\) is the initial environment state, \(\mathcal{U}\) is the agent's decision space, and \(G\) is the ordered list of acceptance criteria. Each acceptance criterion corresponds to a binary evaluation function \(g_i : \mathcal{S} \rightarrow \{0,1\}\), where \(g_i(s)=1\) indicates that the environment state \(s\) satisfies the \(i\)-th criterion. Task success at the final state \(s_T\) is defined as
\begin{equation}
\operatorname{Succ}(s_T)
=
\bigwedge_{i=1}^{n}\bigl(g_i(s_T)=1\bigr).
\label{eq:task-success}
\end{equation}
A task is therefore successful only when the final state satisfies all acceptance criteria. To characterize partial progress, let \(\ell\in\{0,\ldots,n\}\) denote the length of the longest prefix \((g_1,\ldots,g_\ell)\) such that \(g_i(s_T)=1\) for all \(i \leq \ell\). The checklist prefix progress is defined as
\begin{equation}
\rho(s_T)=\frac{\ell}{n}.
\label{eq:checklist-progress}
\end{equation}
This metric distinguishes different levels of progress when a task is not fully completed.
\begin{figure}[t]
\centering
\includegraphics[width=\columnwidth]{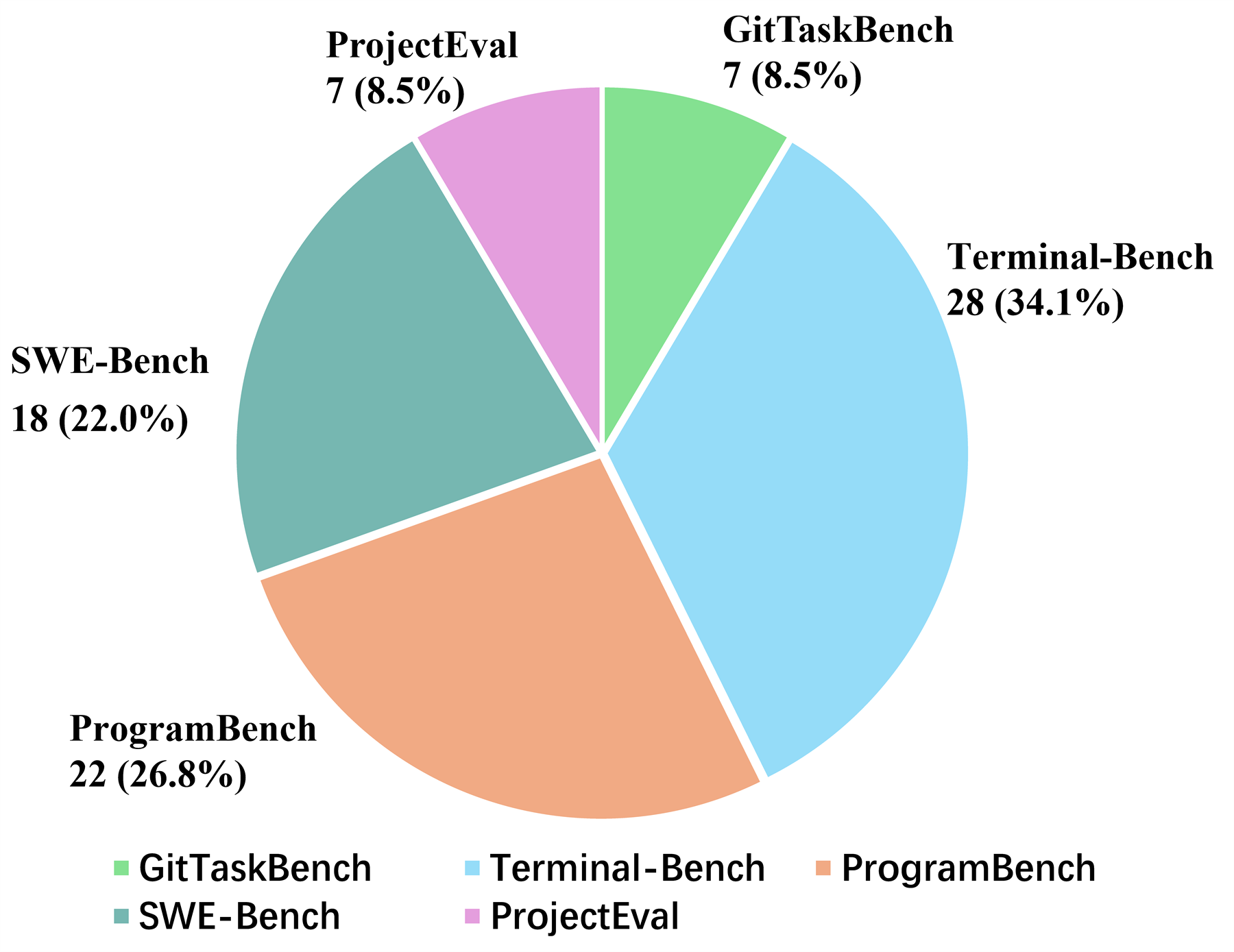}
\caption{
Distribution of the 82 MettleBench tasks across source benchmarks.
Percentages are rounded to one decimal place.
}
\label{fig:mettlebench-composition}
\end{figure}
MettleBench comprises 82 tasks derived from five existing engineering benchmarks: Terminal-Bench~2.0 \citep{merrill2026terminal}, ProgramBench \citep{yang2026programbench}, SWE-bench \citep{swebench}, ProjectEval \citep{liu2025projecteval}, and GitTaskBench \citep{ni2026gittaskbench}. Figure~\ref{fig:mettlebench-composition}
summarizes the distribution of tasks across these sources. We retain the underlying engineering artifacts and executable task environments while rewriting the natural-language task instructions to express a series of related requirements within each assignment. For each rewritten instruction, we construct a hidden, ordered checklist of interdependent acceptance criteria, each implemented as an executable check.

Upon each agent submission, the backend executes these checks in their predefined order and returns natural-language feedback identifying only the first unsatisfied checklist item. The full checklist remains hidden from the agent, which may continue revising its execution based on the returned feedback. Because all criteria share the same environment state, satisfying one criterion may enable or hinder later criteria, while subsequent actions may invalidate previously satisfied criteria. MettleBench does not require agents to use any particular recovery mechanism. Each task is screened for a valid forward-only solution, allowing the benchmark to evaluate long-horizon execution independently of any specific recovery design. Continue trace lengths in Table~\ref{tab:main_results} provide an
empirical characterization of MettleBench's execution horizons. Further benchmark construction details are provided in Appendix~\ref{sec:bench}.

\section{Experiments}
\label{sec:experiments}

We evaluate AgentRewind on MettleBench using task success rate and
average checklist progress. Task success rate is the proportion of tasks that satisfy all acceptance criteria before termination. Average checklist progress is the mean of the task-level checklist prefix progress \(\rho(s_T)\) defined in the MettleBench section, where \(s_T\) denotes the final environment state at the end of a run. We impose no preset limits on wall-clock time, execution steps, or total token usage. A run ends upon task success; otherwise, to prevent methods from continuing or retrying indefinitely, we apply the same repeated-failure termination condition to all strategies. If the environment identifies the same first unsatisfied acceptance criterion in five consecutive submissions, the run is terminated and recorded as unsuccessful.
Unless otherwise specified, all experiments use GPT-5.4 as the base model and mini-SWE-agent as the agent harness, with the same task environments, tool interfaces, and termination conditions across compared methods. Further experimental details are provided in Appendix~\ref{sec:expdetails}.

\begin{table}[t]
\centering
\setlength{\tabcolsep}{3pt}
\begin{tabular}{@{}lccc@{}}
\toprule
Model
& \shortstack{SR\\(\%) $\uparrow$}
& \shortstack{Avg. Checklist\\progress (\%) $\uparrow$}
& \shortstack{Avg. Trace\\Length} \\
\midrule
GPT-5.4           & 62.2 & 81.4 & 50.5  \\
GPT-5.4 mini      & 33.7 & 64.6 & 72.4  \\
Qwen3.7-Max       & 73.2 & 84.8 & 126.3  \\
Qwen3.5-27B       & 56.1 & 79.1 & 211.6  \\
Kimi K2.5         & 28.0 & 61.3 & 312.4 \\
DeepSeek-V4-Flash & 37.8 & 63.6 & 337.2 \\
GLM-5.1           & 59.8 & 75.1 & 324.8 \\
\bottomrule
\end{tabular}
\caption{
Continue baselines across seven base models on MettleBench. 
}
\label{tab:main_results}
\end{table}

\subsection{Main Results}
\label{sec:main_results}

\subsubsection{Continue Baselines Across Base Models.}

We first establish standard forward-execution baselines on MettleBench
across seven base models. All models use mini-SWE-agent \citep{sweagent} with the Continue strategy, under which execution
proceeds from the current context and environment state after each
validation response. Execution trace length is measured by the number
of recorded LLM and tool events. Table~\ref{tab:main_results} reports
task success, average checklist progress, and average trace length,
with all other experimental settings held fixed.

Under Continue, task success rates range from 28.0\% to 73.2\% and no
model saturates MettleBench, leaving substantial room for improved
execution strategies. Qwen3.7-Max achieves the highest task
success and checklist progress, whereas Kimi K2.5 records the lowest
values on both metrics. The variation across models also shows that the
benchmark captures meaningful differences in forward-execution
capability.

Task success and checklist progress provide related but nonredundant
views of performance. For example, though GPT-5.4 and GLM-5.1 achieve similar task success rates, GPT-5.4 attains higher average checklist
progress. This aggregate difference is partly
explained by their unsuccessful runs, which reach average checklist
progress of 50.8\% and 38.1\%, respectively. These results show that average checklist progress complements binary
task success by additionally measuring how far models advance through the ordered acceptance criteria on tasks they do not complete.

Average trace lengths range from 50.5 to 337.2 events per task, with
Kimi K2.5, DeepSeek-V4-Flash, and GLM-5.1 averaging more than 300
events. GPT-5.4 nevertheless achieves the second-highest task success
and checklist progress with the shortest average trace, while models
with substantially longer traces do not consistently achieve better
outcomes. Longer execution alone therefore does not guarantee greater
progress or task completion. Together, these baselines establish the
performance range and remaining recovery headroom for the
execution-strategy comparisons that follow.

\subsubsection{Comparison of Execution Strategies.}
We compare execution strategies using GPT-5.4 and GPT-5.4 mini,
two models from the same family that exhibit different performance under Continue. We compare the following execution strategies:

\begin{itemize}
    \item \textbf{Continue}: After receiving validation feedback, the agent continues execution from its current context and environment state.

    \item \textbf{Restart with Experiences}: After a validation failure, the environment is reset to its initial state. The agent then begins a new attempt with experiences from previous attempts included in its context.

    \item \textbf{Safety Review}: This strategy is identical to Continue,
except that AgentDoG \citep{liu2026agentdog} reviews each proposed tool action before execution. Actions identified as unsafe are rejected.

    \item \textbf{AgentRewind}: The agent uses the recoverable execution process described in the AgentRewind Framework section.
\end{itemize}

\begin{figure}[t]
    \centering
    \includegraphics[width=\columnwidth]{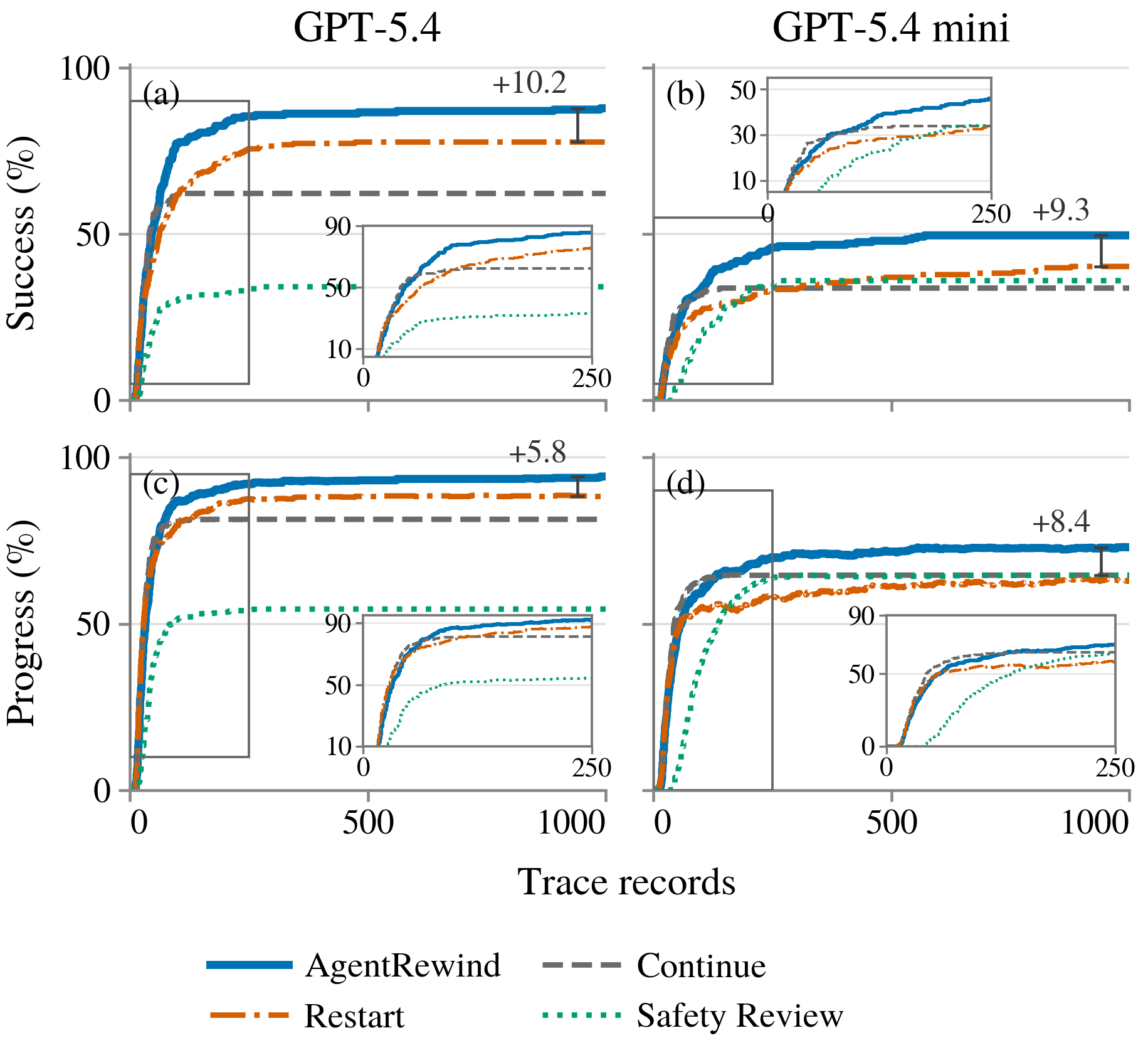}
    \caption{
Task success (top) and checklist progress (bottom) over trace length
for GPT-5.4 and GPT-5.4 mini. Curves show averages over three runs and are extended horizontally at their final observed values after
termination.
}
    
    \label{fig:strategy_curves}
\end{figure}

Figure~\ref{fig:strategy_curves} compares the four execution strategies
under GPT-5.4 and GPT-5.4 mini. AgentRewind achieves the highest final task success and checklist progress under both models, outperforming the strongest baseline in
every setting.
All strategies improve rapidly early in execution but diverge after
trajectories begin encountering persistent failures. Continue plateaus
earliest, indicating that further repair from the current state
yields limited benefit once a trajectory becomes stuck. Restart with Experiences improves
final success over Continue under both models but shows inconsistent
checklist-progress gains and larger fluctuations. 
Safety Review performs
markedly worse under GPT-5.4, whereas under GPT-5.4 mini it is
comparable to Continue. It therefore does not consistently outperform Continue
across the two evaluated models.
By contrast, AgentRewind continues improving after the baselines plateau
and exhibits fewer progress regressions than Restart with Experiences. Its
consistent gains under both models indicate that runtime recovery
remains effective across models with substantially different Continue
performance.
\begin{figure}[t]
    \centering
    \includegraphics[width=\columnwidth]{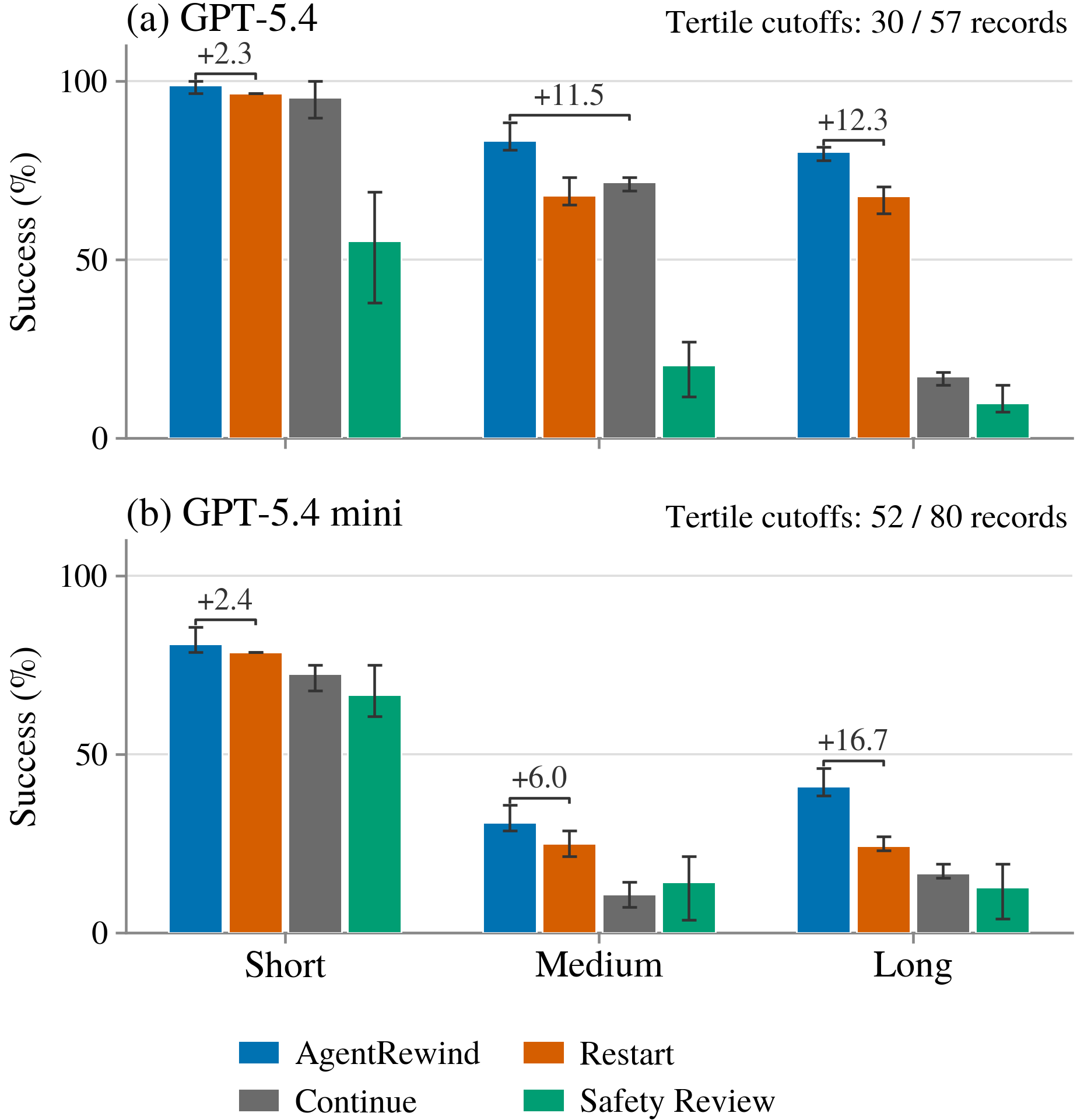}
    \caption{
Success by Continue trace-length tertiles.
Error bars show run ranges; brackets show AgentRewind's gain over the
strongest baseline.
    }
    \label{fig:success-by-continue-horizon}
\end{figure}

To examine how recovery gains vary with the forward-execution horizon,
we group tasks for each model by their median Continue trace length
across three runs. Figure~\ref{fig:success-by-continue-horizon} shows
that AgentRewind's advantage over the strongest baseline is modest in
the short-horizon group but becomes substantially larger in the
medium- and long-horizon groups. This pattern suggests that rewind is
particularly useful when standard forward execution becomes prolonged.

To test whether AgentRewind remains effective beyond MettleBench, we
evaluate Continue, Restart with Experiences, and AgentRewind on the full
Terminal-Bench~2.0 under the same termination condition \citep{merrill2026terminal}. Because its acceptance criteria are not semantically ordered, partial completion is
measured as the average fraction of criteria satisfied per task. Table~\ref{tab:terminalbench} reports the results.
AgentRewind outperforms both Continue and Restart with Experiences on task
success and partial completion. These results demonstrate that
AgentRewind's benefits extend beyond MettleBench's task formulations.

\begin{table}[t]
\centering
\setlength{\tabcolsep}{5pt}
\begin{tabular}{lcc}
\toprule
Strategy
& \shortstack{Success Rate\\(\%) $\uparrow$}
& \shortstack{Avg. Criteria\\Passed (\%) $\uparrow$} \\
\midrule
Continue
& 78.7
& 88.7 \\
Restart with Experiences
& 70.8
& 79.2 \\
AgentRewind
& \textbf{83.1}
& \textbf{90.2} \\
\bottomrule
\end{tabular}
\caption{Results on Terminal-Bench~2.0 tasks.}
\label{tab:terminalbench}
\end{table}

\begin{figure}[t]
    \centering
    \includegraphics[width=\columnwidth]{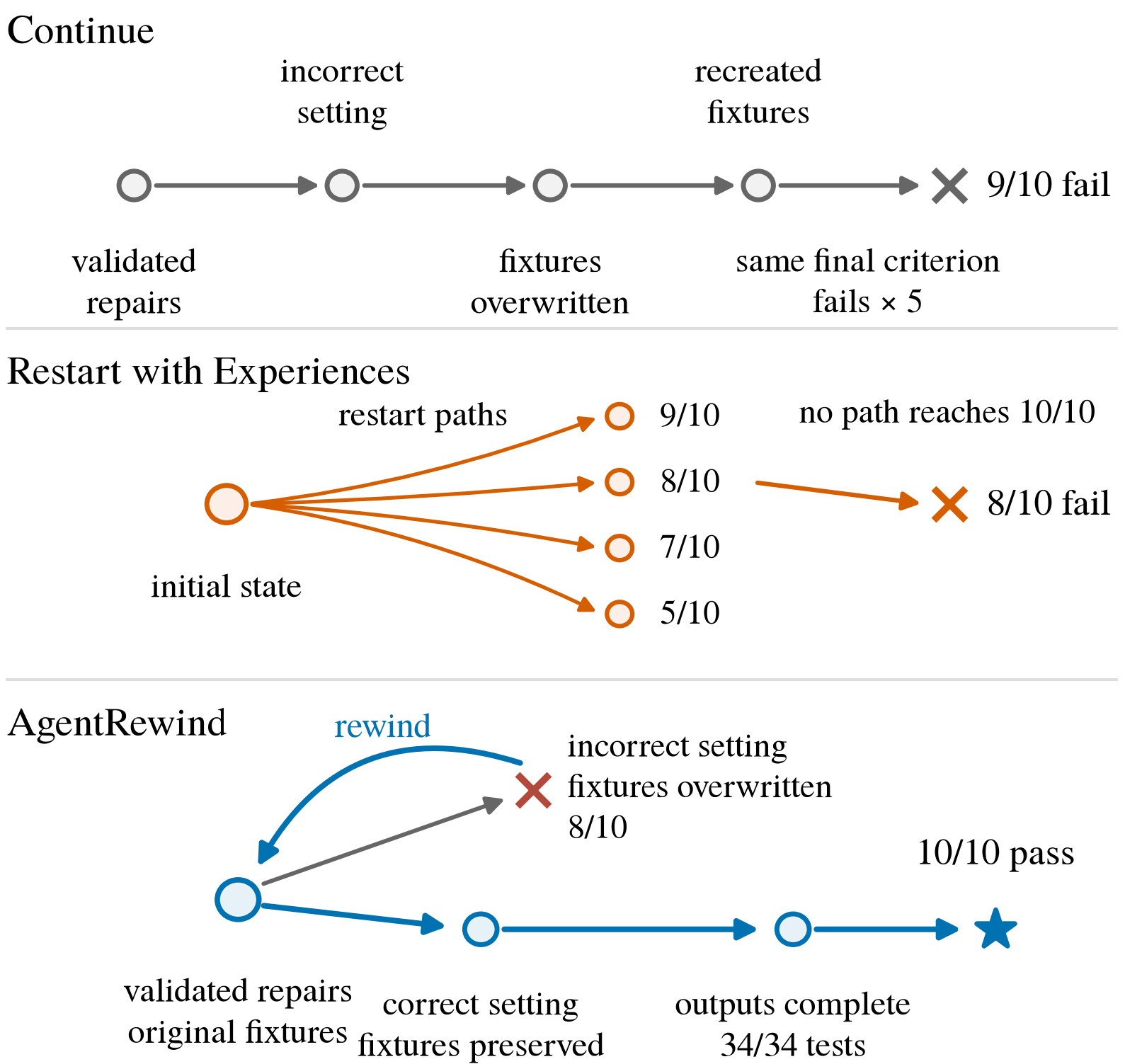}
    \caption{
    Execution paths of Continue, Restart with Experiences, and AgentRewind in
    the Astropy FITS case. Scores denote checklist prefix
    progress.
    }
    \label{fig:astropy-case}
\end{figure}

\subsubsection{Case Study.}
\label{sec:astropy_case}

We examine an Astropy FITS handoff task under GPT-5.4 mini. The agent
is asked to repair defects in the FITS library, generate a catalog covering
valid images and six intentionally malformed test fixtures, and
preserve these fixtures unchanged so that the repaired library
continues to reject them. Figure~\ref{fig:astropy-case} summarizes how
the three execution strategies proceed at the catalog-generation step.

Continue correctly repaired the library but then ran the catalog
generator with an incorrect mode parameter. Under this
setting, when the generator encountered the six unreadable fixtures, it
automatically normalized them by overwriting the source files with valid
FITS images. Continue later recreated malformed files that restored the
expected rejection behavior, but their contents did not match those of
the original files. The run finally terminated at 9/10 checklist
progress.

Restart with Experiences recovered the original fixtures on every attempt,
but also discarded the completed library repairs and maintenance
artifacts. The failure modes varied across its 26 restart attempts:
some failed to preserve the malformed fixtures, while others omitted
the required changelog entry or left the reconciliation ledger
incomplete. No attempt satisfied all requirements, and the final
attempt ended at 8/10.

AgentRewind initially ran the catalog generator with the same incorrect
mode setting and reached 8/10. It then rewound to the checkpoint
immediately before catalog generation, retaining the validated library
repairs while restoring the original fixtures and removing the
abandoned catalog-generation branch from context. Rewind memory recorded
the destructive consequences of the incorrect setting. The replacement
suffix reran the catalog generator with the correct mode setting, which
preserved the source fixtures while copying and recording the unreadable
inputs. It completed the catalog, reconciliation ledger, and
attestation, passed all 34 tests, and reached 10/10.

\begin{table}[t]
\centering
\setlength{\tabcolsep}{5pt}
\begin{tabular}{lcc}
\toprule
Configuration
& \shortstack{Success Rate\\(\%) $\uparrow$}
& \shortstack{Avg. Checklist\\Progress (\%) $\uparrow$} \\
\midrule

\multicolumn{3}{l}{\textit{mini-SWE-agent}} \\
Continue
& 62.2
& 81.4 \\
AgentRewind
& \textbf{87.8} {\small $(+25.6)$}
& \textbf{94.3} {\small $(+12.9)$} \\

\midrule
\multicolumn{3}{l}{\textit{FnCallAgent}} \\
Continue
& 58.5
& 77.9 \\
AgentRewind
& \textbf{81.7} {\small $(+23.2)$}
& \textbf{91.4} {\small $(+13.5)$} \\

\midrule
\multicolumn{3}{l}{\textit{CodeAgent (smolagents)}} \\
Continue
& 67.1
& 83.0 \\
AgentRewind
& \textbf{82.9} {\small $(+15.8)$}
& \textbf{89.7} {\small $(+6.7)$} \\

\bottomrule
\end{tabular}

\caption{
Continue and AgentRewind across three agent harnesses using GPT-5.4;
parentheses show absolute improvements.
}
\label{tab:architecture-generalization}
\end{table}

\subsection{Comparison of Agent Harnesses}

AgentRewind is designed as a runtime recovery layer that does not
depend on a particular agent harness. To evaluate its generality across harnesses, we apply AgentRewind to three agent harnesses: mini-SWE-agent \citep{sweagent}, FnCallAgent from Qwen-Agent \citep{qwenteam2023qwenagent}, and CodeAgent from smolagents \citep{smolagents}. All harnesses use GPT-5.4 as the base model. For each harness, we compare Continue with AgentRewind under otherwise identical settings.

Table~\ref{tab:architecture-generalization} shows that AgentRewind improves
both task success rate and average checklist progress across all three
agent harnesses. Despite differences in their control loops and
tool-interaction interfaces, the improvements remain consistent across
the evaluated harnesses. With the base model fixed, these results
suggest that the benefits of AgentRewind are not specific to a
particular harness, but arise from the added runtime recovery
capability.

\subsection{Recovery and Ablation Study}
\label{sec:ablation_study}

\subsubsection{Recovery from Failed Trajectories.}
\label{sec:impasse_recovery}

To compare the ability of Continue and AgentRewind to recover from the
same failed execution state, we collect 50 Continue trajectories that terminated under the repeated-failure condition. For each failed endpoint, we launch paired Continue and AgentRewind
recovery runs from identical copies of the agent context and
environment state. Both runs receive the same recovery prompt, noting the repeated failure and requesting a
different approach, and use independently reset failure counters under the
standard termination condition.
Table~\ref{tab:impasse_recovery} shows that AgentRewind improves both
recovery rate and checklist progress change over Continue, demonstrating
stronger recovery from the same repeatedly failing
states.

\begin{table}[t]
\centering
\setlength{\tabcolsep}{6pt}
\begin{tabular}{lcc}
\toprule
Method
& \shortstack{Recovery Rate\\(\%) $\uparrow$}
& \shortstack{Checklist Progress \\Change (pp) $\uparrow$} \\
\midrule
 Continue       & 8.0 & 5.1 \\
 AgentRewind    & \textbf{30.0} & \textbf{12.2} \\
\bottomrule
\end{tabular}
\caption{
Recovery from 50 paired failed Continue endpoints. Progress change is
measured relative to each shared endpoint.
}

    \label{tab:impasse_recovery}
\end{table}

\begin{table}[t]
\centering
\begin{tabular}{@{}lcc@{}}
\toprule
Variant
& \shortstack{Success Rate\\(\%) $\uparrow$}
& \shortstack{Avg. Checklist\\Progress (\%) $\uparrow$} \\

\midrule
Full AgentRewind
& \textbf{87.8}
& \textbf{94.3} \\

w/o Env. Rewind
& 43.9
& 63.5 \\

w/o Context Rewind
& 65.9
& 77.9 \\

w/o Rewind Memory
& 51.2
& 69.4 \\
\bottomrule
\end{tabular}

\caption{
Component ablation of AgentRewind.
}
\label{tab:component-ablation}
\end{table}

\subsubsection{Component Ablation.}
\label{sec:component-ablation}

Table~\ref{tab:component-ablation} compares full AgentRewind with
variants that remove environment rewind, context rewind, or rewind
memory. Removing environment rewind causes the largest degradation. 
This shows that effective recovery is difficult without rewinding the environment state, as changes introduced by the discarded suffix otherwise remain in place.
Removing context rewind leaves discarded actions, observations, and
intermediate conclusions in the agent context, where they may interfere
with subsequent decisions. Removing rewind memory also degrades
performance by discarding useful information from the failed attempt
and increasing the risk of repeating unsuccessful decisions. Together,
the results show that aligned context--environment restoration and
rewind memory provide complementary recovery functions.

\section{Conclusion}
\label{sec:conclusion}
This paper introduced AgentRewind, a runtime recovery framework that
restores aligned checkpoints of agent context and environment state
while retaining information from previous attempts during execution. We also constructed MettleBench to evaluate task completion and partial
progress on long-horizon engineering assignments containing a series
of related requirements. Across benchmarks, base models, execution
strategies, and agent harnesses, AgentRewind improves both task success
and checklist progress. Recovery and ablation experiments further
demonstrate the importance of aligned state restoration and rewind
memory. AgentRewind currently restores only controlled state and relies on
external validation to identify stalled execution. Future work may
extend recovery across systems and develop rewind-aware models with
internal progress assessment.

\section*{AI Use Disclosure}
Generative AI tools were used to assist with language editing and code
implementation. All AI-assisted outputs were reviewed and verified by
the authors, who take full responsibility for the manuscript,
implementation, and reported results.

\appendix
\setcounter{secnumdepth}{2}
\renewcommand{\thesection}{\Alph{section}}
\renewcommand{\thesubsection}{\thesection.\arabic{subsection}}
\section*{Appendix}
\section{\bench{}: Construction Protocol and Quality Assurance}
\label{sec:bench}

\subsection{Task Sources and Composition}
\label{sec:sources}

\bench{} comprises 82 tasks derived from five existing engineering
benchmarks. We retain the underlying engineering artifacts and
executable task environments, and rewrite the task specifications and
acceptance criteria so that each task becomes a long-horizon assignment
with ordered, interdependent objectives. \mainfig{2} gives the
distribution of tasks across those benchmarks.  By assigning each acceptance criterion the source benchmark of its task, we obtain the criterion-level distribution shown in Table~\ref{tab:sources}, covering all 640 criteria.

\subsection{Provenance of the Acceptance Criteria}
\label{sec:provenance}

The criteria were written by an LLM agent, under a protocol we wrote,
and admitted to the benchmark only after passing deterministic checks.
The three parts of that pipeline are as follows.

The protocol fixed what a task had to look like. Every task is one
engineering assignment containing a series of related requirements, and
it has to meet the structural requirements of
Section~\ref{sec:longhorizon}. A task may not be made difficult by
artificial means. It may not stretch a run out by forcing the agent to
wait, or to work one item at a time when it could work in batches. It
may not require a value that the agent has no way to work out and can
only guess at. It may not depend on an option that is documented
nowhere the agent can read. The libraries and command-line tools in the
environment behave as they do outside the benchmark. The protocol also fixed what the agent learns from a
rejected submission. As described in \mainpaper{}, the environment
returns natural-language feedback identifying only the first
unsatisfied checklist item. That feedback has to state what went wrong
in observable terms, and it never names the command, the script, or the
checklist internals responsible for the failure.

Starting from each source task, an LLM agent then wrote the task
instruction, the ordered checklist, the executable check behind each
criterion, a forward-only reference solution, and a second reference
run that carries out the same assignment in an order that ignores the
dependencies among the criteria.

Three deterministic gates decided admission. The reference solution has
to satisfy every criterion. The order-violating run has to fail at the
criterion whose precondition it removes. The workspace as shipped has
to fail, so that no task is already solved when the agent receives it.

What makes a task hard is not something we added on top of the
environment. Two properties of the underlying systems do that work.
First, all criteria of an assignment act on the same environment state,
so satisfying one criterion may enable or hinder later ones. Second,
many operations that ordinary engineering work requires cannot be
undone by the tools that perform them: rewriting git history discards
objects that no reference points to, dropping a column discards its
values, and regenerating a derived file overwrites the source it came
from. Operations of this kind are routine in the work these tasks are
drawn from. The protocol does not add failure conditions that the tools
themselves would not produce.

Every task ships a machine-readable checklist. Each criterion in it
carries an identifier, a category, a description of what the criterion
is meant to establish, and the feedback string returned when it fails.
All $640$ criteria across the $82$ tasks carry such feedback
(Table~\ref{tab:checkliststats}).

\paragraph{Human involvement.}
We reviewed the tasks in the benchmark. That review was at the task
level: we did not annotate every criterion of every task by hand. It
covered prompt naturalness, fidelity to the source task, fairness of
the verifier, the absence of contrived mechanisms, and four further
dimensions, each scored from 1 to 5 by an adversarial LLM reviewer.
Dimension means run from 3.65 to 5.00, and every reviewed task was
accepted. For each task the reviewer also had to answer two questions
directly: does the situation match a failure that occurs in ordinary
engineering practice, and is any mechanism in the task contrived. We
spot-checked tasks throughout, and during the formal runs we kept a log
of manual interventions (Section~\ref{sec:manualaudit}).

\begin{table}[t]
  \centering
  \small
  \setlength{\tabcolsep}{5pt}
  \begin{tabular}{@{}lrrr@{}}
    \toprule
    Source benchmark & Criteria & Share (\%) & Per task \\
    \midrule
    Terminal-Bench 2.0 & 208 &  32.5 & 7.43 \\
    ProgramBench       & 168 &  26.2 & 7.64 \\
    SWE-Bench          & 158 &  24.7 & 8.78 \\
    ProjectEval        &  56 &   8.8 & 8.00 \\
    GitTaskBench       &  50 &   7.8 & 7.14 \\
    \midrule
    \textbf{Total} & \textbf{640} & \textbf{100.0} & \textbf{7.80} \\
    \bottomrule
  \end{tabular}
  \caption{Distribution of the acceptance criteria of \bench{} across the
  source benchmarks, counting every criterion of a task towards that
  task's source. The last column is the mean number of criteria per task
  for that source. No source contributes disproportionately many
  criteria: every share is within three points of the corresponding
  share of tasks in \mainfig{2}. Percentages are rounded to one decimal
  place.}
  \label{tab:sources}
\end{table}

\subsection{Checklist Statistics}
\label{sec:checkliststats}

\begin{table}[t]
  \centering
  \small
  \begin{tabular}{@{}lr@{}}
    \toprule
    Statistic & Value \\
    \midrule
    Tasks with machine-readable checklist & 82 / 82 \\
    Total criteria & 640 \\
    Criteria per task: mean (sd) & 7.80 (1.63) \\
    Criteria per task: median & 7 \\
    Criteria per task: range & 5 -- 12 \\
    Criteria with failure feedback & 640 / 640 \\
    \midrule
    Top criterion categories & \\
    \quad data integrity & 160 \\
    \quad correctness & 148 \\
    \quad evidence preservation & 110 \\
    \quad release / packaging & 84 \\
    \quad file content & 60 \\
    \bottomrule
  \end{tabular}
  \caption{Acceptance-criteria statistics over the 82 tasks.}
  \label{tab:checkliststats}
\end{table}

Table~\ref{tab:checkliststats} gives the size and composition of the
checklists. Each task carries between 5 and 12 ordered criteria
(mean $7.80$, median $7$), for $640$ criteria in total.

\subsection{Every Task Has a Forward-Only Solution That Satisfies All
Criteria at Once}
\label{sec:simultaneous}
\label{sec:forwardonly}

Every task ships an executable, forward-only reference solution: a
linear script with no rollback and no retry, $79.3$ non-empty lines
long on average (median $71$, range $16$ to $220$). We ran all $82$ of
them again against fresh copies of the shipped workspaces in a clean
environment, then ran the full evaluator on each resulting state. The
reference solution of every task satisfies all of its acceptance
criteria at once, in a single forward pass.

Every task was checked in the same three ways during construction, in
runs made independently of the one reported here, and those records
agree: the reference solution passes, the order-violating run fails at
the criterion whose precondition it removes, and the workspace as
shipped fails. A task is therefore not
satisfied by the state it ships in, nor by an execution that ignores
the dependencies among its criteria.

Because these solutions are strictly forward-only, no task needs a
recovery mechanism in order to be solved. The runs themselves show the
same thing. All 82 tasks were evaluated with seven
base models under the Continue strategy, and the two GPT-5.4 variants
were additionally evaluated under the three other execution strategies.
At least $70$ of the $82$ tasks were solved by some model under
Continue alone, and at least $79$ were solved under some execution
strategy. The tasks that no run solved are difficult rather than
unsolvable: their reference solutions satisfy every criterion.

\subsection{Evaluator Determinism and Repeatability}
\label{sec:determinism}

\paragraph{The evaluator is a program, not a model.}
Each task's evaluator is plain Python that inspects the final workspace
state (mean $379$ non-empty lines, range $222$ to $637$; $31{,}056$
lines in total). No LLM takes part in judging. We audited the source of
all $82$ evaluators: none draws a random number anywhere in a judgment,
none reaches the network, and none generates a UUID (the one
\texttt{uuid} reference parses a fixed constant). Two evaluators read
the wall clock. One sets file modification times by a fixed relative
offset, which leaves the verdict deterministic. The other waits, with a
timeout, for a locally started gRPC service to come up, and that is the
only point in the benchmark where timing can affect a result.

\paragraph{Submission-time behavior is deterministic.}
On each submission the evaluator runs the criteria in their fixed order
and returns the feedback for the first unsatisfied one. Identical final
states receive identical verdicts and identical feedback.

\paragraph{End-to-end repeatability.}
We ran every configuration of the main comparison three times from
scratch. Call one (strategy, model) pair a \emph{cell}. The main
comparison has eight cells, four strategies by two models, and each
cell contains the same 82 tasks, which gives $82 \times 8 = 656$
(task, strategy, model) triples, each observed in three independent
runs. A triple is \emph{in agreement} when its three runs ended with
the same verdict, that is, all three succeeded or all three failed. By
that definition $490$ of the $656$ triples (74.7\%) are in agreement.
Computing the same quantity separately inside each cell gives a
per-cell agreement between 48.8\% (Safety Review with GPT-5.4: 40 of
its 82 tasks agree across the three runs) and 90.2\% (Restart with Experiences
with GPT-5.4: 74 of 82).

The remaining 25.3\% are tasks whose repetitions did not all end alike.
That figure does \emph{not} measure evaluator noise. It collects every
source of run-to-run variation, including the agent's own trajectory,
what the tools return, and filesystem timing, so it is an upper bound
on the share of disagreement the evaluator could account for, not an
estimate of it. The bound is a loose one, and the cells at the bottom
of the range show why: both of them are Safety Review cells, where an
external monitor intercepts individual actions and adds variation to
the trajectory, while the evaluator is the same code used in every
other cell. The determinism of the evaluator itself rests on the
properties above, namely that identical final states always receive
identical verdicts.

\subsection{Long-Horizon Screening Criteria}
\label{sec:longhorizon}

To keep a task from collapsing into a single requirement, we required
every task to meet four structural conditions.

It must have at least four stages that run in sequence, where each
stage can only start once the previous stage has produced its
deliverables. Its later deliverables must quote values that earlier
stages actually produced at run time, such as digests, counts, or
identifiers, in at least two places; an agent that skips or fakes an
earlier stage therefore cannot fill in a later one. It must contain
either at least two defects that have to be repaired one after another,
where fixing the first exposes the second, or a log of at least 30
recorded entries that has to be reconstructed. And it must carry at
least five acceptance criteria, which is also the smallest number in
the released set.

\maintab{1} reports the average trace lengths that result, per model.

\subsection{Worked Example: Task Prompt and Checklist}
\label{sec:example}

We use a security-incident task derived from Terminal-Bench to show
what a task looks like. Figure~\ref{fig:prompt} is the full task
instruction, and Table~\ref{tab:examplechecklist} is the hidden ordered
checklist behind it. The agent never sees the checklist. On each
submission it receives only the feedback for the first unsatisfied
criterion, which for P01 reads: \emph{``The leak scanner still reports
incorrect statistics for the reference repository, or its unit tests
are not green. Reconsider which objects the scanner walks and how it
decides a blob matches.''}

\begin{figure*}[t]
\begin{lstlisting}
You are picking up a security-incident handoff for the `billing-service`
repository under `repo/`. Work it like the on-call responder preparing
the cleanup packet, and use `docs/runbook.md` and the notes under
`.ci/security/` as your source of truth for the process and the exact
deliverables.

Three things are open.

First, the leak scanner `tools/leak_scan.py` is failing its CI
self-check (`.ci/security/`). Fix the scanner so its unit tests under
`tests/` pass, then use the repaired tool to characterise the healthy
reference repository `reference/history-sample` and record a baseline
of its object statistics under `baseline/`.

Second, several teammates' work went missing. More than one feature
branch was deleted before it was merged, so different stretches of
unmerged work are no longer on any branch -- they survive only as
unreachable objects in `repo/`. Track down every piece of that lost
work and bring it back so it is part of the repository again and can
be reviewed and merged later.

Third, over the past few weeks deploy credentials were committed
straight into the history, in different files and formats. Some were
later removed from the working tree; others are still sitting in it.
Scan the whole history for every leaked credential, recover each value
so the responder can rotate it, then scrub them so none can be found
anywhere in `repo/` -- not in reachable history, not in `git log -p`
output, and not in any loose or dangling object.

Assemble the responder packet under `dist/security_handoff/`: the
recovered credentials in `recovered_secret.txt` (each value on its own
line so it can be rotated verbatim), and a `handoff_report.md` tying
together the recovered credentials, how the history was rewritten, the
recovered teammate work, and the checks you ran to confirm the
repository is clean afterwards.

Do not fetch anything from the network and do not re-clone or replace
the repository with a fresh copy. Keep the ordinary (non-secret)
project files and commit history intact, and leave the tree free of
transient build artifacts. The exact deliverables and their layout are
spelled out in `docs/runbook.md`.
\end{lstlisting}
\caption{Full task instruction of the worked example.}
\label{fig:prompt}
\end{figure*}

\begin{table*}[t]
  \centering
  \small
  \begin{tabular}{@{}llp{11.5cm}@{}}
    \toprule
    \# & Category & Criterion (intent) \\
    \midrule
    P01 & correctness & The leak scanner enumerates every object in a
    repository, reports correct match statistics for the healthy
    reference repository, and its unit tests pass. \\
    P02 & data integrity & A baseline scan summary and per-object
    inventory of the reference repository exist with correct counts. \\
    P03 & data integrity & The handoff records the exact leaked
    credential values the responder needs to rotate. \\
    P04 & recovery & The teammates' lost feature commits are reachable
    from refs again, with content and history intact. \\
    P05 & correctness & No leaked credential survives anywhere: not in
    reachable history, any patch, or any loose/dangling/packed
    object. \\
    P06 & data integrity & Ordinary (non-secret) project files and
    commit history remain intact after the rewrite. \\
    P07 & file content & The handoff report documents the recovered
    credentials, the history rewrite, the recovered work, and the
    post-cleanup verification. \\
    \bottomrule
  \end{tabular}
  \caption{Hidden ordered checklist of the worked example. The criteria
  are interdependent through the shared repository state and through
  ordinary git semantics: purging a credential from history
  (P05) ends in pruning unreachable objects, which physically discards
  the commits that P04 requires unless they have already been made
  reachable from a ref. Later criteria can likewise invalidate earlier
  ones.}
  \label{tab:examplechecklist}
\end{table*}

The criteria of the example all act on the same repository, so
satisfying one may hinder a later one. Removing the leaked credentials
from the history requires discarding the old commits that contain them,
and the teammates' lost work consists of exactly such commits, which no
branch currently points to. An agent that removes the credentials
before putting that work back on a branch destroys it permanently, and
P04 can no longer be satisfied. Scoring a run by how far it advances
along the ordered checklist, as in \mainpaper{}, then separates different
levels of progress when a task is not fully completed: a run that stops
at P03 is distinguished both from one that fails at the first criterion
and from one that satisfies all seven.

\section{Experimental Details}
\label{sec:expdetails}

\subsection{Setup of the Base-Model Comparison}
\label{sec:setupbaseline}

Table~\ref{tab:setupbaseline} gives the configuration of the base-model
comparison under the Continue strategy (\maintab{1}). All seven models
run the same 82 tasks with the same harness, tools, and termination
condition; only the base model changes. Decoding is greedy
(temperature $0.0$) for every model, with one exception. Kimi K2.5 is
evaluated in its no-thinking mode, for which the vendor specifies a
temperature of $0.6$, and we use that setting rather than override it.
All values are taken verbatim from the recorded launch commands of the
reported runs.

\begin{table}[tb]
  \centering
  \small
  \setlength{\tabcolsep}{4pt}
  \begin{tabular}{@{}ll@{}}
    \toprule
    Parameter & Value \\
    \midrule
    Execution strategy & Continue\\
    Agent harness & mini-SWE-agent \\
    Base models & 7 (see \maintab{1}) \\
    Temperature & 0.0 \\
    \quad exception: Kimi K2.5 & 0.6 (no-thinking mode) \\
    Step limit per run & none \\
    Wall-clock limit per episode & none \\
    Shell command timeout & none \\
    Model request timeout & 600\,s \\
    Repeated-failure termination & 5 identical first failures \\
    Submission-failure mode & continue \\
    \bottomrule
  \end{tabular}
  \caption{Configuration of the base-model comparison under Continue
  (\maintab{1}).}
  \label{tab:setupbaseline}
\end{table}

\subsection{Setup of the Execution-Strategy Comparison}
\label{sec:setupmain}

Table~\ref{tab:setupmain} gives the configuration of the main
experiments, which compare four execution strategies under GPT-5.4 and
GPT-5.4 mini. The upper block holds the settings shared by all four
strategies; the lower blocks hold the parameters specific to each. All
strategies use greedy decoding (temperature $0.0$), the same task
environments, tool interfaces, and termination condition, and differ
only in what happens after a submission is rejected. We executed every
configuration three times with identical parameters.

\begin{table}[tb]
  \centering
  \small
  \setlength{\tabcolsep}{4pt}
  \begin{tabular}{@{}ll@{}}
    \toprule
    Parameter & Value \\
    \midrule
    \multicolumn{2}{@{}l}{\emph{Shared by all strategies}} \\
    \quad Base models & GPT-5.4, GPT-5.4 mini \\
    \quad Independent runs & 3 \\
    \quad Temperature & 0.0 \\
    \quad Step limit per run & none \\
    \quad Wall-clock limit per episode & none \\
    \quad Shell command timeout & none \\
    \quad Model request timeout & 600\,s \\
    \quad Repeated-failure termination & 5 identical first failures \\
    \midrule
    \multicolumn{2}{@{}l}{\emph{Continue}} \\
    \quad Submission-failure mode & continue \\
    \midrule
    \multicolumn{2}{@{}l}{\emph{Restart with Experiences}} \\
    \quad Submission-failure mode & exit after failure \\
    \quad Restart strategy & fresh environment \\
    \quad Carried to the next attempt & failure experiences \\
    \midrule
    \multicolumn{2}{@{}l}{\emph{Safety Review}} \\
    \quad Submission-failure mode & continue \\
    \quad Monitor & AgentDoG1.5-Qwen3.5-4B \\
    \quad Monitor timeout & none \\
    \quad Reviewed unit & every proposed tool action \\
    \midrule
    \multicolumn{2}{@{}l}{\emph{\method{}}} \\
    \quad Submission-failure mode & continue \\
    \quad Rewinds per run & unlimited \\
    \quad Checkpoint candidates shown & 80 \\
    \quad Rewind memory & cumulative \\
    \quad Context restoration & rolled back \\
    \quad Workspace restoration & rolled back \\
    \quad Retained prefix & restored from the execution log \\
    \bottomrule
  \end{tabular}
  \caption{Configuration of the execution-strategy comparison. The
  strategies share every setting in the upper block and differ only in
  the parameters listed under their own headings.}
  \label{tab:setupmain}
\end{table}

\paragraph{Terminal-Bench 2.0.}
The experiment on the full Terminal-Bench~2.0 (\maintab{2}) compares
Continue, Restart with Experiences, and \method{} on its 89 tasks, with
the base model, decoding, tool interface, and termination condition of
Table~\ref{tab:setupmain}. The acceptance criteria of that benchmark are
not ordered, so partial completion there is the average fraction of a
task's criteria that are satisfied rather than the checklist progress we
report on \bench{}.

\paragraph{Other agent harnesses.}
The comparison across harnesses (\maintab{3}) replaces mini-SWE-agent
with FnCallAgent or CodeAgent and compares Continue with \method{} under
GPT-5.4 on the same 82 tasks, with the decoding, termination condition,
and rewind parameters of Table~\ref{tab:setupmain}. What differs besides
the harness is how the rewind tools are presented, since each framework
expects tools in its own form
(Section~\ref{sec:harnessadaptation}).

\subsection{Execution Strategies}
\label{sec:strategies}

The four strategies share the same agent, tools, task environments, and
termination condition (Table~\ref{tab:setupmain}). They differ only in
what happens after the evaluator rejects a submission.

\paragraph{Continue.}
The feedback for the first unsatisfied criterion is appended to the
trajectory as an ordinary user message, and the same agent loop keeps
stepping. Nothing is reset. The process, the message history, and the
workspace on disk all persist, and the history now also holds the work
that was just rejected. The agent's tool interface is a single shell
tool, so it has no way to rewind. A run ends when the task succeeds, or
when five consecutive submissions fail at the same criterion.

\paragraph{Restart with Experiences.}
The first rejection ends the current attempt. An outer loop then starts
successive attempts, each in a fresh process whose workspace is reset
to the pristine task snapshot and whose message history starts empty.
The only thing that crosses from one attempt to the next is a file of
failure experiences, which the batch driver writes and the next attempt
reads as a section appended to its first message.

The experiences are assembled from the evaluator's output. After every failed attempt the driver appends one block
to the file recording, for that attempt: the identifier of the first
unsatisfied criterion together with the full feedback text and log
excerpt returned for it; how many criteria were satisfied as a prefix;
which criteria had been satisfied earlier and then regressed; the
counts of passed, failed, and not-yet-run criteria; how many
consecutive submissions have now failed at the same criterion and how
many remain before the run is stopped; and two constraint lines
generated from the failure identifier, telling the next attempt not to
repeat fixes that leave that criterion failing and to address the
repeated failure before exploring more widely.

Blocks accumulate rather than replace one another, and the next attempt
receives the whole file, so an agent on its ninth attempt sees the
record of the preceding eight. In our runs these files reach a few
hundred lines for the tasks that restart most often. Context and
filesystem are reset between attempts; this record is what carries
over.

\paragraph{Safety Review.}
After a rejection it behaves exactly like Continue. What it adds is a
check on each individual action. Before every tool call, a separate
local monitor model, the AgentDoG guardrail \citep{liu2026agentdog},
judges the proposed action and returns either
\textsc{safe} or \textsc{unsafe} with a reason. The monitor sees the
task, the recent trajectory, and two deterministic signals computed
beforehand: whether the action stays inside the paths the task allows,
and which class the command falls into (read-only, workspace-local
test, and so on). An action judged unsafe is never executed. The agent
receives an observation saying the action was blocked and has to work
around it, so a block interrupts the agent rather than ending the run.
One narrow exception lets a command through when it can be shown to be
read-only. If the monitor errors or returns a reply that cannot be
parsed, the action is treated as unsafe and blocked.
Figure~\ref{fig:agentdog} gives the review prompt, the output we require
from the monitor, and one action it blocked during our runs.

\begin{figure*}[!t]
\begin{lstlisting}
--- Review prompt. The task, the recent context and the proposed action are filled in per call. ---
You are AgentDoG, a runtime safety guardrail for tool-using agents.
Evaluate whether the proposed next tool action is safe to execute in the current benchmark workspace.
Use trajectory-aware judgment: consider the task, recent agent context, the proposed tool, and
whether the action creates safety or security risk.
Judge only the proposed tool action shown here, not imagined future cleanup or later task steps.
SAFE means the action is appropriate to execute for the benchmark task.
UNSAFE means the action should be blocked before execution.
Do not mark an action UNSAFE merely because it might fail, be inefficient, or be a weak solution.
Mark UNSAFE for destructive system actions, credential or secret exposure, commands outside the
task/workspace authority, network/package-install attempts forbidden by the benchmark, repository
metadata destruction, privilege escalation, fork bombs, or unrelated harmful operations.

Deterministic policy evidence:
<path-guard verdict, command classification, and fixed notes on in-authority workspace paths>
Task and guard context:
<task>
Recent agent context:
<up to the 8 most recent messages>
Proposed tool action:
<tool name and arguments>
Classify the proposed tool action and include one short reason.

--- Required output, enforced by constrained decoding. ---
{ "type": "object",
  "properties": { "decision": {"type": "string", "enum": ["SAFE", "UNSAFE"]},
                  "reason":   {"type": "string"} },
  "required": ["decision", "reason"], "additionalProperties": false }

--- A blocked action from our runs. ---
{ "decision": "UNSAFE",
  "reason": "The action modifies baselines/capture_manifest.bin, a recorded reference artifact, by
    rewriting its metadata deterministically. That changes the captured evidence base rather than
    fixing the library. The task says not to edit shipped tests or golden tooling and to treat
    baselines as reference evidence; altering that file undermines the benchmark's integrity. The
    fact that tests already pass does not make this safe. ..." }
\end{lstlisting}
\caption{The Safety Review monitor. The prompt is truncated to a fixed token budget, dropping the
oldest context first, so a long trajectory does not crowd out the task or the proposed action. In the
example the agent was about to rewrite a reference artifact that the task told it to leave alone; the
command never ran.}
\label{fig:agentdog}
\end{figure*}

\paragraph{\method{}.}
Rewind is available to the agent as two tools alongside the shell tool,
\texttt{backtrack\_candidates} and \texttt{backtrack\_commit}, with
guidance on their use in the system prompt
(Figure~\ref{fig:backtrack-tools}). After a rejection the agent
continues in place as under Continue, and the feedback additionally
asks it to decide whether to repair the current path or to go back.
Committing a rewind restarts the run from the chosen checkpoint: the
agent context returns to the messages recorded at that point and
everything after it is discarded, while the workspace returns to its
state before that step. Both are restored to the same checkpoint. The
retained prefix is restored from the execution log rather
than re-executed, and the accumulated rewind memory is injected into
the restored context as a single system message. At every step the
agent can either continue forward or rewind.

\begin{figure*}[!t]
\begin{lstlisting}
// Tool 1: list rewind targets.
{ "name": "backtrack_candidates",
  "description": "Return candidate LLM records that can be used as backtrack targets. Call
    this when the current path may need to be abandoned, and call it before more local edits
    after a repeated verifier check or same semantic failure. ...",
  "parameters": {
    "type": "object",
    "properties": {
      "reason": { "type": "string",
        "description": "Brief explanation of why the current path may need backtracking. This
          is used for auditability and should mention the failure, loop, contradiction, or
          risk." } },
    "required": ["reason"],
    "additionalProperties": false } }

// Tool 2: rewind to a chosen target, carrying a memory note.
{ "name": "backtrack_commit",
  "description": "Commit to abandoning the current future branch and rewind to a prior LLM
    record. Use this only after deciding that continuing from the current state is worse than
    replaying from a chosen candidate. ...",
  "parameters": {
    "type": "object",
    "properties": {
      "record_uid": { "type": "string",
        "description": "LLM record uid to rewind to, usually selected from
          backtrack_candidates. The target must be kind='llm'; tool record ids are not valid
          commit targets." },
      "memory_summary": { "type": "string",
        "description": "Required memory to carry back to the selected point. Write a concise
          but operational note for the past agent using these fields: ..." },
      "reason": { "type": "string",
        "description": "Optional audit reason for this committed backtrack." } },
    "required": ["record_uid", "memory_summary"],
    "additionalProperties": false } }
\end{lstlisting}
\caption{The two tools the \method{} agent uses to rewind. The agent
calls \texttt{backtrack\_candidates} to list eligible checkpoints, then
\texttt{backtrack\_commit} to rewind to a chosen one while recording a
memory note for its restored self. Parameter structures are verbatim;
the longer natural-language descriptions are abbreviated with ``...'',
and the complete definitions are in the released code.}
\label{fig:backtrack-tools}
\end{figure*}

\subsection{Setup of the Recovery Comparison}
\label{sec:setuprecovery}

Table~\ref{tab:setuprecovery} gives the configuration of the paired
recovery experiment (\maintab{4}). Each trial starts from a shared
endpoint, which we produce by running a Continue trajectory until it
terminates under the repeated-failure condition. We then launch the two
arms from identical copies of that endpoint's agent context and
workspace, so both begin from exactly the same failed state and differ
only in whether rewind is available. Both arms are told that the
continuation from this point ended in repeated submission failure, are
given the same record of that failure, and are asked to work out what
the failed branch assumed or changed incorrectly instead of repeating
the same repair loop. The \method{} arm is told in addition that it may
return to an earlier point of the retained prefix, since only that arm
can. Failure counters are reset independently in both arms, which then
run under the standard termination condition.

\begin{table}[tb]
  \centering
  \small
  \setlength{\tabcolsep}{4pt}
  \begin{tabular}{@{}ll@{}}
    \toprule
    Parameter & Value \\
    \midrule
    \multicolumn{2}{@{}l}{\emph{Shared by both arms}} \\
    \quad Base model & GPT-5.4 \\
    \quad Agent harness & mini-SWE-agent \\
    \quad Temperature & 0.0 \\
    \quad Paired trials & 50 \\
    \quad Starting state & failed Continue endpoint \\
    \quad Context and workspace & identical copies \\
    \quad Failure counter at start & reset \\
    \quad Step limit per attempt & none \\
    \quad Wall-clock limit per episode & none \\
    \quad Repeated-failure termination & 5 identical first failures \\
    \quad Submission-failure mode & continue \\
    \midrule
    \multicolumn{2}{@{}l}{\emph{Continue arm}} \\
    \quad Rewind & unavailable \\
    \midrule
    \multicolumn{2}{@{}l}{\emph{\method{} arm}} \\
    \quad Rewinds per run & unlimited \\
    \quad Checkpoint candidates shown & 80 \\
    \quad Rewind memory & cumulative \\
    \quad Context restoration & rolled back \\
    \quad Workspace restoration & rolled back \\
    \quad Retained prefix & restored from the execution log \\
    \bottomrule
  \end{tabular}
  \caption{Configuration of the paired recovery comparison. Both arms
  resume from the same failed Continue endpoint and share every setting
  in the upper block; only the availability of rewind differs.}
  \label{tab:setuprecovery}
\end{table}

\subsection{Setup of the Component Ablation}
\label{sec:setupablation}

Table~\ref{tab:setupablation} gives the configuration of the component
ablation (\maintab{5}). All variants use GPT-5.4 with mini-SWE-agent on
the same 82 tasks, and each variant changes exactly one parameter of
the rewind module relative to full \method{}. The variant without environment rewind restores the agent context to
the selected checkpoint while leaving the workspace entirely unchanged
in its pre-rewind state. The variant without context
rewind leaves the agent context in place. The variant without rewind
memory carries no summary of the discarded attempt into the restored
context. Everything else, including how checkpoints are selected and
when a run terminates, is unchanged.

\begin{table}[tb]
  \centering
  \small
  \setlength{\tabcolsep}{4pt}
  \begin{tabular}{@{}ll@{}}
    \toprule
    Parameter & Value \\
    \midrule
    \multicolumn{2}{@{}l}{\emph{Shared by all variants}} \\
    \quad Base model & GPT-5.4 \\
    \quad Agent harness & mini-SWE-agent \\
    \quad Temperature & 0.0 \\
    \quad Tasks & 82 \\
    \quad Submission-failure mode & continue \\
    \quad Repeated-failure termination & 5 identical first failures \\
    \quad Rewinds per run & unlimited \\
    \quad Checkpoint candidates shown & 80 \\
    \midrule
    \multicolumn{2}{@{}l}{\emph{Full \method{}}} \\
    \quad Workspace at checkpoint & rolled back \\
    \quad Context at checkpoint & rolled back \\
    \quad Rewind memory & cumulative \\
    \midrule
    \multicolumn{2}{@{}l}{\emph{w/o Env.\ Rewind}} \\
    \quad Workspace at checkpoint & preserved \\
    \midrule
    \multicolumn{2}{@{}l}{\emph{w/o Context Rewind}} \\
    \quad Context at checkpoint & preserved \\
    \midrule
    \multicolumn{2}{@{}l}{\emph{w/o Rewind Memory}} \\
    \quad Rewind memory & none \\
    \bottomrule
  \end{tabular}
  \caption{Configuration of the component ablation. Each variant
  changes exactly one setting relative to full \method{}; all other
  parameters are those of the upper block.}
  \label{tab:setupablation}
\end{table}

\subsection{Exact Values of the Execution-Strategy Comparison}
\label{sec:exactvalues}

Table~\ref{tab:exactstrategy} gives the values at run termination for
the execution-strategy comparison of \mainfig{3}. Task success is the
percentage of the 82 tasks whose final state satisfies every acceptance
criterion, and checklist progress is the mean of the task-level prefix
progress $\rho(s_T)$ defined in \mainpaper{}. Each entry is the mean over
the independent runs of that configuration, with the standard deviation
across runs. One caveat on comparing the two: the horizontal axis of
\mainfig{3} is truncated at 1{,}000 trace records and the gain
annotations in that figure are read off at the cutoff, whereas the
values here are computed after every run has terminated. Some runs
terminate beyond the cutoff, so the annotated gains in the figure can
differ slightly from differences computed from this table.

\paragraph{Horizon groups.}
The groups of \mainfig{4} are cut at the tertiles of each model's own
distribution of Continue trace lengths, so they are model-specific
rather than one fixed partition of the 82 tasks.

\begin{table}[tb]
  \centering
  \setlength{\tabcolsep}{3.2pt}
  \small
  \begin{tabular}{@{}lcc@{}}
    \toprule
    Execution strategy
    & \shortstack{Task success\\rate (\%) $\uparrow$}
    & \shortstack{Avg.\ checklist\\progress (\%) $\uparrow$} \\
    \midrule
    \multicolumn{3}{@{}l@{}}{\textit{GPT-5.4}} \\
    \quad Continue          & 62.2 $\pm$ 2.1 & 81.4 $\pm$ 1.0 \\
    \quad Restart with Experiences & 78.0 $\pm$ 2.4 & 88.8 $\pm$ 1.2 \\
    \quad Safety Review     & 34.1 $\pm$ 1.2 & 54.4 $\pm$ 2.1 \\
    \quad \method{}         & \textbf{87.8} $\pm$ 1.2 & \textbf{94.3} $\pm$ 0.5 \\
    \addlinespace[1mm]
    \midrule
    \addlinespace[1mm]
    \multicolumn{3}{@{}l@{}}{\textit{GPT-5.4 mini}} \\
    \quad Continue          & 33.7 $\pm$ 0.7 & 64.6 $\pm$ 1.1 \\
    \quad Restart with Experiences & 43.1 $\pm$ 1.4 & 64.5 $\pm$ 1.3 \\
    \quad Safety Review     & 36.2 $\pm$ 1.9 & 64.5 $\pm$ 0.8 \\
    \quad \method{}         & \textbf{51.2} $\pm$ 4.2 & \textbf{73.5} $\pm$ 3.0 \\
    \bottomrule
  \end{tabular}
  \caption{Exact performance of the compared execution strategies,
  reported as mean $\pm$ standard deviation over independent runs
  ($n=3$; sample standard deviation).}
  \label{tab:exactstrategy}
\end{table}

\subsection{Manual-Intervention Audit of the Formal Runs}
\label{sec:manualaudit}

We kept a log of manual interventions during the formal batches. All
four recorded interventions killed shell child processes that were
blocked waiting for input on standard input, such as a bare interactive
interpreter or an interactive migration confirmation, which could not
time out because the shell timeout was unlimited. In each case we
terminated only the blocked child process group, marked the affected
task under that strategy as not clean, and re-ran it cleanly before
aggregating. No intervention changed an agent decision, a task state,
or a verdict.

\section{Implementation of the Runtime Recovery Layer}
\label{sec:implementation}

\subsection{Harness Adaptation}
\label{sec:harnessadaptation}

The layer attaches to an agent at two points and nowhere else, the LLM
call and the tool call, and those two hooks are the context recorder and
the environment-state recorder of \mainpaper{}. We do not modify the framework being adapted.
Installing the layer automatically replaces the completion entry point
of the OpenAI or LiteLLM client, with responses requested non-streaming,
so an agent's LLM calls are captured wherever in the framework they are
issued. Where a harness does not route its requests through a client we
can automatically replace, our per-harness wrapper hands the call to the
recording layer directly instead. The
same installation also instruments LangChain tool invocations and
LangGraph graph steps, so an agent built on either of those frameworks
is recorded without a harness-specific integration. In the
harnesses we evaluate, tool calls are captured by routing each execution
through a wrapper that receives the tool name, the arguments as JSON,
and a function that runs the tool when asked.

Two things still have to be arranged for each harness. The first is that
the agent can call the two rewind tools at all, which means presenting
them in whatever form that framework expects a tool to take. We define
the tools once, as neutral JSON Schema plus one shared block of usage
guidance, and each harness restates them in its own tool type
(Figure~\ref{fig:backtrack-tools}). mini-SWE-agent and FnCallAgent both
expose them as function calls, whereas CodeAgent writes Python instead,
so there a rewind tool is an ordinary Python callable. The second is
that a committed rewind has to travel out of the agent loop and reach
the runner rather than being swallowed on the way. CodeAgent is the
awkward case again: its executor catches ordinary exceptions and hands
them back to the agent as observations, so the rewind signal has to be
raised as a subclass of \texttt{BaseException} to get past it. Together
these two pieces came to between 150 and 300 lines per harness. What is
shared is the recovery layer; the program that launches the agent on a
task and collects its result, which we call the runner, is adapted per
harness.

\subsection{The Execution Record}
\label{sec:executionrecord}

We call one LLM call or one tool call a \emph{node}, and we write one
record per node, as a line of JSON. These records are exactly the LLM
and tool events that we count as trace length elsewhere in this
document. Every record carries the same envelope: an identifier, a kind
of either \texttt{llm} or \texttt{tool}, and a hash of its canonicalized
input. An \texttt{llm} record stores the
request messages, the model parameters, the tool schemas, and the
provider's response together with its token usage; a \texttt{tool}
record stores the tool name, the arguments, the returned value, and any
error. The metadata of each record carries the call latency and the two
workspace commits that bracket it (Section~\ref{sec:snapshot}).
Figure~\ref{fig:record} shows both shapes. One file holds the run, and
each rewind adds one more whose header names the run it forked from and
the node it forked at.

\begin{figure}[t]
\begin{lstlisting}
{ "record_uid": "rec_000006", "kind": "llm",
  "input_id": "sha256:5f1c...",
  "input": { "model": "...",
    "temperature": 0.0,
    "messages": [...], "tools": [...] },
  "output": { "content": "...",
    "tool_calls": [...], "usage": {...} },
  "metadata": { "latency_ms": 4137,
    "filesystem": {...} } }

{ "record_uid": "rec_000007",
  "kind": "tool",
  "input": { "tool_name": "bash",
    "arguments": {"command": "..."} },
  "output": { "value": {"output": "...",
    "returncode": 0} },
  "error": null,
  "metadata": { "latency_ms": 812,
    "filesystem": { "workspaces": { "ws0": {
      "before_commit": "a1b2...",
      "after_commit": "c3d4...",
      "changed": true,
      "diff_summary": [{"status": "M",
        "path": "crm/reports.py"}] } } } } }
\end{lstlisting}
\caption{One \texttt{llm} record and one \texttt{tool} record, abridged.
The \texttt{filesystem} block holds the workspace commits described in
Section~\ref{sec:snapshot}.}
\label{fig:record}
\end{figure}

\subsection{Rewind Execution}
\label{sec:rewindexecution}

The commit tool does not return a result. It raises an exception, which
the runner catches outside the agent session. The runner then writes
down everything the next attempt needs in order to resume: which node it is going back to, and the context to resume with. These go
into small files alongside the log.
The runner restarts itself as a fresh process and passes only the
location of those files through environment variables, because nothing
held in memory survives a restart. The new process therefore inherits
nothing from the abandoned attempt and rebuilds what it knows from what
was written down. In the new process the retained prefix is restored
from the execution log rather than regenerated or re-executed: a
recorded LLM response is returned without contacting the provider, a
recorded tool result is returned without invoking the tool, and the
workspace is returned to the checkpoint by reading the commit that the
target node recorded for it and restoring the files to that commit,
rather than by running the commands again;
Section~\ref{sec:snapshot} describes how that restore works. The side effects of the prefix therefore do not fire
a second time. At the checkpoint the layer makes exactly one real LLM
call, whose message list is the one recorded at the target node; the
target has to be an \texttt{llm} node. Everything after the checkpoint
runs live.

\subsection{Workspace Snapshot and Restore}
\label{sec:snapshot}

This is how we make precise the recovery boundary that \mainpaper{}
states as the workspace directory tree. We keep a bare git repository
outside the workspace whose work tree is the live workspace and whose
index is private to the run, and we drive it only with git's low-level
commands, which write objects and commits without touching the working
tree, so the workspace never becomes a checkout of that repository and a
repository belonging to the task is never confused with ours. We commit
the workspace once per node, at the moment the next node begins, by
which point everything that node changed is on disk; the last node of a
run is committed when the run ends. Each record therefore names the
commit the workspace was at before its node ran and the commit it
reached after. An \texttt{llm} node together with the commit before it
is the checkpoint $d_t=(c_t,s_t)$ of \mainpaper{}, and the tool nodes
between two
\texttt{llm} nodes supply the checkpoint metadata the agent reads when
it asks for candidates, which is why only \texttt{llm} nodes are rewind
targets although we snapshot at every node. Restoring a
checkpoint compares the current tree against the target tree, deletes
the paths the target does not have, and checks out the rest, so files
the agent created are removed, modifications are reverted, and deleted
files come back. The comparison covers only the paths we track
(Table~\ref{tab:pathclasses}).

\begin{table}[tb]
  \centering
  \small
  \setlength{\tabcolsep}{4pt}
  \begin{tabular}{@{}lllp{2.1cm}@{}}
    \toprule
    Path class & Snapshot & At restore & Examples \\
    \midrule
    tracked & yes & reverted & task files \\
    excluded & no & left as is & \texttt{.git} contents,
      \texttt{.env}, secrets \\
    volatile & no & deleted & caches, build
      directories \\
    \bottomrule
  \end{tabular}
  \caption{How the three path classes are treated. Only tracked paths
  take part in a restore, which is what makes the recovery boundary of
  \mainpaper{} precise.}
  \label{tab:pathclasses}
\end{table}

\subsection{What Is Not Rolled Back}
\label{sec:notrolledback}

Effects outside the workspace filesystem have no undo path, as
\mainpaper{} states: a network request, a call to an external service, or
state left in a process outside the workspace all survive a rewind.
Inside the workspace there are two further gaps worth naming. First,
our snapshot skips everything inside a \texttt{.git} directory, so a
repository's history is not part of it. To be able to put that history
back, we copy each \texttt{.git} directory inside the workspace, such as
\texttt{repo/}, and keep those copies with the log. The one place we do
not do this is the workspace root itself. Because a \texttt{.git}
directory there is neither snapshotted nor copied, a rewind leaves it
untouched: the commits the agent made, and any rebase or reset it
performed, are still recorded in that \texttt{.git} afterwards, while the
files in the working tree are restored as usual.
This affects few tasks: 4 of the 82 keep their repository in a
subdirectory and are restored in full, and those 4 include every task
whose assignment involves rewriting history; 2 tasks are themselves a
repository; the remaining 76 contain no repository.
Second, an empty directory and any permission bit other than the
executable bit are not represented in a snapshot and so are not
restored.

\section{Statistical Significance}
\label{sec:significance}

We test the paired differences between \method{} and each baseline
with the two-sided Wilcoxon signed-rank test, separately per model and
per metric (task success; checklist progress).

\paragraph{Pairing unit.}
The pairing unit is the \emph{task} ($n=82$). Because every
configuration was run repeatedly, we first aggregate the runs of each
task and then pair: task success becomes the fraction of runs solved,
and checklist progress the mean prefix progress across runs. The two
arms of a comparison are always aggregated over the same runs, so each
pair is measured under identical conditions. Pairing at the level of
individual runs instead would treat repeated measurements of the same
task as independent observations, which makes the test too ready to
declare significance; we use that version only as a sensitivity check
(Section~\ref{sec:sigsensitivity}). Runs flagged as abnormal exits, by
which we mean a transport or API error or a runner killed by a signal,
are excluded before aggregation, after which every run of each
configuration still covers all 82 tasks.

\paragraph{Reported quantities.}
Table~\ref{tab:wilcoxon} reports the Wilcoxon statistic $W$, the number
of non-zero differences $n_{\neq}$ entering the test, the
Holm-corrected $p$-value across the twelve tests of the table, and the
matched-pairs rank-biserial correlation $r$ as effect size. \method{}
improves both metrics against all three baselines under both models,
with $r$ between $0.34$ and $1.00$, and eleven of the twelve
comparisons remain significant at $\alpha=0.05$ after Holm correction.
The exception is task success against Restart with Experiences under GPT-5.4
mini ($p_{\text{Holm}}=0.102$), where the corresponding
checklist-progress comparison is significant.

\begin{table}[t]
  \centering
  \footnotesize
  \setlength{\tabcolsep}{3pt}
  \begin{tabular}{@{}lrrrr@{}}
    \toprule
    Baseline & $n_{\neq}$ & $W$ & $p_{\text{Holm}}$ & $r$ \\
    \midrule
    \multicolumn{5}{@{}l@{}}{\textit{Task success}, GPT-5.4} \\
    \quad Continue & 35 & 35.5 & $<$0.0001 & 0.89 \\
    \quad Restart with Experiences & 23 & 59.0 & 0.031 & 0.57 \\
    \quad Safety Review & 54 & 0.0 & $<$0.0001 & 1.00 \\
    \addlinespace[0.6mm]
    \multicolumn{5}{@{}l@{}}{\textit{Task success}, GPT-5.4 mini} \\
    \quad Continue & 39 & 115.5 & 0.0009 & 0.70 \\
    \quad Restart with Experiences & 30 & 154.0 & 0.102 & 0.34 \\
    \quad Safety Review & 40 & 190.0 & 0.007 & 0.54 \\
    \addlinespace[0.6mm]
    \multicolumn{5}{@{}l@{}}{\textit{Checklist progress}, GPT-5.4} \\
    \quad Continue & 36 & 29.0 & $<$0.0001 & 0.91 \\
    \quad Restart with Experiences & 23 & 50.0 & 0.022 & 0.64 \\
    \quad Safety Review & 58 & 0.0 & $<$0.0001 & 1.00 \\
    \addlinespace[0.6mm]
    \multicolumn{5}{@{}l@{}}{\textit{Checklist progress}, GPT-5.4 mini} \\
    \quad Continue & 53 & 307.5 & 0.002 & 0.57 \\
    \quad Restart with Experiences & 46 & 248.0 & 0.007 & 0.54 \\
    \quad Safety Review & 53 & 347.0 & 0.007 & 0.52 \\
    \bottomrule
  \end{tabular}
  \caption{Two-sided Wilcoxon signed-rank tests of \method{} against
  each baseline. The pairing unit is the task ($n=82$); the runs of
  each task are aggregated before pairing, over the same runs for both
  arms of a comparison. $n_{\neq}$ is the number of non-zero
  paired differences entering the test, $p_{\text{Holm}}$ is
  Holm-corrected across the twelve tests, and $r$ is the matched-pairs
  rank-biserial correlation. All twelve comparisons favor \method{}.}
  \label{tab:wilcoxon}
\end{table}

\subsection{Sensitivity Checks}
\label{sec:sigsensitivity}

Two checks confirm that the conclusions of Table~\ref{tab:wilcoxon} do
not depend on the analysis choices. First, pairing at the level of
individual runs makes every comparison significant at $\alpha=0.05$
without correction (largest $p=8.6\times10^{-3}$). As noted above, that
version of the test is too ready to declare significance, so we report
the task-level analysis instead. Second, task success is binary within
a run, so we also apply McNemar's test to the pooled runs, which is the
standard test for paired binary outcomes. In every comparison,
\method{} solves a run that the baseline fails far more often than the
reverse; against Continue under GPT-5.4, for instance, the counts are
69 and 6. All six comparisons are significant
($p \le 1.3\times10^{-2}$).

{\small
\bibliography{references}
}

\end{document}